\documentclass{article}

\usepackage[preprint, nonatbib]{neurips_2021}

\usepackage[utf8]{inputenc} 
\usepackage[T1]{fontenc}    
\usepackage{hyperref}       
\usepackage{url}            
\usepackage{booktabs}       
\usepackage{amsfonts}       
\usepackage{nicefrac}       
\usepackage{microtype}      
\usepackage{xcolor}         
\usepackage{amsmath} 
\usepackage[capitalize]{cleveref}
\usepackage[detect-all,per-mode=symbol]{siunitx}
\usepackage{booktabs}
\usepackage{multirow}
\usepackage{pgfplots}
\newcommand{\argmax}{\operatornamewithlimits{arg\,max}}

\usepgfplotslibrary{groupplots}
\usetikzlibrary{positioning}
\pgfplotsset{compat=newest}
\pgfplotsset{every axis legend/.append style={%
cells={anchor=west}}
}
\pgfplotsset{every axis/.append style={
                    label style={font=\footnotesize},
					tick label style={font=\footnotesize},
					legend style={font=\footnotesize}
                    }}
\usetikzlibrary{arrows}
\tikzset{>=stealth'}

\usepackage[style=ieee, maxcitenames=2, mincitenames=1, backend=bibtex]{biblatex}
\AtEveryBibitem{
 	\clearfield{url}
 	\clearfield{doi}
\ifentrytype{inproceedings}{
 	\clearlist{address}
 	\clearlist{publisher}
 	\clearname{editor}
 	\clearlist{organization}
 	\clearfield{pages}
 	\clearlist{location}
	\clearfield{volume}
	\clearfield{series}
 }{}
 }

\AtBeginBibliography{\small}

\title{Coordinated Reinforcement Learning \\ for Optimizing Mobile Networks}

\author{%
  Maxime Bouton,
  Hasan Farooq,
  Julien Forgeat,
  Shruti Bothe,
  Meral Shirazipour,
  Per Karlsson
  \\
  Ericsson Research, Santa Clara CA, USA \\
  \texttt{\{maxime.bouton, hasan.farooq, julien.forgeat,
  shruti.bothe\}@ericsson.com}
}

\begin{document}

\maketitle

\begin{abstract}
	Mobile networks are composed of many base stations and for each of them many parameters must be optimized to provide good services.
	Automatically and dynamically optimizing all these entities is challenging as they are sensitive to variations in the environment and can affect each other through interferences.
	Reinforcement learning (RL) algorithms are good candidates to automatically learn base station configuration strategies from incoming data but they are often hard to scale to many agents.
	In this work, we demonstrate how to use coordination graphs and reinforcement learning in a complex application involving hundreds of cooperating agents.
	We show how mobile networks can be modeled using coordination graphs and how network optimization problems can be solved efficiently using multi-agent reinforcement learning.
	The graph structure occurs naturally from expert knowledge about the network and allows to explicitly learn coordinating behaviors between the antennas through edge value functions represented by neural networks.
	We show empirically that coordinated reinforcement learning outperforms other methods.
	The use of local RL updates and parameter sharing can handle a large number of agents without sacrificing coordination which makes it well suited to optimize the ever denser networks brought by 5G and beyond.
\end{abstract}

\section{Introduction}

One of the major factors influencing the quality of experience in mobile networks is the configuration of base station antennas, notably the tilt angle.
Incorrectly configured base stations antennas in a mobile network could interfere with neighboring antennas and deteriorate the signal of users in a cell that would otherwise have good coverage. To appropriately configure a network before deployment, engineers have to anticipate many possible traffic conditions as well as possible sources of interference from the environment and the network itself.
With a constantly growing demand in high quality services, an increasing network complexity, and highly dynamic environments, relying on human interventions to update network configurations leads to a suboptimal use of the network resources and is often very costly.

Instead, one could automate the optimization procedure under the context of Self-Organizing Networks (SONs)~\cite{aliu2013survey}. Existing approaches for network optimization rely on hand-engineered strategies which are suboptimal and hard to scale~\cite{eckhardt2011vertical,eisenblatter2008capacity,saeed2012controlling}.
Methods relying on mathematical models~\cite{partov2015utility} or reinforcement learning (RL) are also used for network optimization~\cite{farooq2019ai,dandanov2017dynamic,balevi2019online,shafin2020self,galindo2010distributed,vanella2020offpolicy}. They are more robust and principled. However, due to the large scale and the complex interactions between network entities they often consider them as independent agents and do not leverage the benefit of cooperation.
The resulting solution is suboptimal despite its scalability.
On the other hand optimizing all agents together can yield better solutions at the expense of scalability.

In this work, we propose to derive a coordinated RL approach for dynamic network optimization taking into account local interactions between the base stations.
We propose a generic approach to model cellular networks as coordination graphs.
The graph representation, provides a support for encoding prior knowledge on a network deployment and can be obtained from standard network specifications.
Leveraging the structure of the graph allows using a scalable distributed optimization to find an approximately optimal joint antenna configuration through message passing.
In order to dynamically adapt to the environment, we intertwine the optimization procedure with local RL updates that learn value functions for each edge of the graph, taking advantage of the network topology.
To improve the sample efficiency of the RL algorithm and enabling its scalability to hundreds of learning agents we share network parameters across the graph edges which allows using data from many pairs of agents to train the same network.
The proposed method can scale to very large networks and is well suited to handle the expected increase in density of the networks brought by 5G and beyond.
We empirically show that coordinated RL scales to more than two hundred agents and, when augmented with parameter sharing, consistently outperforms other multi-agent RL algorithms on an antenna tilt control problem.
This work demonstrates how multi-agent RL algorithms can be applied beyond simple games, to a variety of large scale network optimization problems involving coordination between agents.

\section{Related Work}

Most prior works on network optimization often address the problem by controlling one antenna independently of each other thus ignoring potential coordination benefits.
They rely on rule-based methods~\cite{eckhardt2011vertical,eisenblatter2008capacity} or optimization techniques using mathematical models~\cite{partov2015utility}. These methods are often suboptimal and hard to scale to many agents~\cite{saeed2012controlling} because the search space increases exponentially with the number of agents. Instead, one can focus on reinforcement learning to automatically learn control strategies from incoming network data.

When using Reinforcement learning, the problem of coordination still arises. Considering each antenna as independent learning agents have been used in the past to address the problem of optimizing mobile networks~\cite{farooq2019ai,balevi2019online,shafin2020self,vanella2020offpolicy,nasir2019multiagent}, hence failing to capture phenomena like interference. Learning algorithms leveraging coordination can use a centralized controller~\cite{dandanov2017dynamic,calabrese2018learning} which does not scale to a large number of agents. In contrast, our approach benefits from recent advances in multi-agent reinforcement learning to enable scalable coordination achieving an approximately optimal joint optimization of the whole network.

Prior work using the mean-field multi-agent RL algorithm~\cite{yang2018meanfield} has shown promising results for antenna tuning problems~\cite{balevi2019online}. This method requires aggregating the influence of all the neighbors through a resulting action. This action is not easy to estimate in practice and does not distinguish which neighbor has the most influence. Our work demonstrates that multi-agent RL techniques through value function factorizations~\cite{oroojlooyjadid2019review} is a scalable approach for learning cooperative behavior with hundreds of controlled agents. A key enabler is to model interactions using a coordination graph. Contrary to other techniques like QMIX~\cite{rashid2018qmix}, it allows encoding prior knowledge through the graph structure. We show empirically that the learned edge value functions can be interpreted.

The procedure of using coordination graphs for RL was first introduced by \citeauthor{guestrin2002coordinated}\cite{guestrin2002coordinated}.
Update equations for learning the edge value functions are very well described and justified by \citeauthor{kok2006collaborative} for tabular representations of the value function \cite{kok2005maxplus}.
Our proposed algorithm combines coordinated RL with neural network approximation and parameter sharing.
The core of the algorithm is similar to \citeauthor{boehmer2020deep}~\cite{boehmer2020deep} while some aspects are simplified in order to allow for a fully distributed implementation in a real mobile networks.
Our method does not make any requirements on the graph structure and we propose an approach to derive coordination graphs from existing mobile network deployments.

\section{Collaborative Multi-agent Reinforcement Learning}

Collaborative sequential decision making problems can be modeled by a multi-agent Markov decision process where several decision agents are trying to maximize a common reward function.
Formally, it is defined by a tuple $(\mathcal{N}, \mathcal{S}, \mathcal{A}, T, R, \gamma)$ where $\mathcal{N}$ is a set of $n$ agents, $\mathcal{S}=\mathcal{S}_1\times\ldots\times\mathcal{S}_n$ a joint state space, $\mathcal{A} =\mathcal{A}_1\times\ldots\times\mathcal{A}_n$ a joint action space, $T$ a transition function, $R$ a global reward function and $\gamma \in [0,1)$ a discount factor.
We make an additional assumption that the reward function can be additively decomposed into local reward signals: $R(\mathbf{s}, \mathbf{a}) = \sum_{i=1}^n r_i(\mathbf{s}, \mathbf{a})$. Note that the individual reward still depends on the joint state and action as agents might influence each other's performance.

In this work, we seek to find a joint policy mapping a joint state to an action that maximizes the accumulated global reward. We define the value function associated to a policy $\pi$ as ${Q^\pi(\mathbf{s}, \mathbf{a}) = E[\sum_{t=0}^\infty\gamma^t\sum_{i=1}^n r_{i,t} | \mathbf{s_0}=\mathbf{s}, \mathbf{a_0} = \mathbf{a}]}$. Given a joint value function, a joint policy is given by ${\pi(\mathbf{s}) = \argmax_\mathbf{a} Q(\mathbf{s}, \mathbf{a})}$.
The value function can be approximated by a parameteric model learned by minimizing the following Bellman loss: ${J(\theta) = \mathbb{E}_{\mathbf{s}'}[(r + \gamma \max_{\mathbf{a}'}Q(\mathbf{s}', \mathbf{a}';\theta) - Q(\mathbf{s}, \mathbf{a}; \theta))^2]}$ where $\theta$ represents learned parameters and $(\mathbf{s}, \mathbf{a}, r, \mathbf{s}')$ is an experience tuple obtained by interacting with the environment.

The size of the joint state, and action spaces grows exponentially with the number of agents, making the value function difficult to learn and represent even with powerful function approximators such as neural networks.
Computing the argmax to extract the action from the value function suffers from combinatorial explosion.
Previous works in MARL often attempts to learn value function decomposition such as in VDN (linear) or QMIX~\cite{sunehag2018value,rashid2018qmix} to make this problem tractable.
If knowledge about dependencies between agents is available, such decomposition can directly be used through a coordination graph


A coordination graph $\mathcal{G}$ is defined by a set of vertices $\mathcal{V}$ and a set of undirected edges $\mathcal{E}$.
Each vertex corresponds to an agent.
Each edge $(i,j)\in\mathcal{E}$ is associated to a value function $Q_{ij} : \mathcal{S}_i\times\mathcal{S}_j\times\mathcal{A}_i\times\mathcal{A}_j\rightarrow\mathbb{R}$ representing the pairwise value function of agent $i$ and $j$.
The topology of the graph allows for capturing coordination behavior between connected agents.
For each edge, $(i,j)$, we initialize a value function $Q_{ij}$ as a function parameterized by $\theta$.
With knowledge of the graph topology, the joint value function can then be expressed as follows:
\begin{equation}
	Q(\mathbf{s}, \mathbf{a}) = \sum_{(i,j)\in\mathcal{E}} Q_{ij}(s_i, s_j, a_i, a_j; \theta)
\end{equation}
To address high dimensional and continuous observation spaces, we represent the value functions by neural networks, $Q_{ij}(s_i, s_j, a_i, a_j;\theta)$ for all $(i,j) \in \mathcal{E}$.
The parameters are shared across all edges.
To learn these pairwise value functions agents will keep interacting with the environment in a loop alternating between action selection and learning.

\textbf{Action selection.} The action selection procedure consists of solving the following problem:
\begin{equation}
	\label{eq:inference}
	\mathbf{a^*} = \argmax_{(a_1, \ldots, a_n)} \sum_{(i,j)\in\mathcal{E}} Q_{ij}(s_i, s_j, a_i, a_j)
\end{equation}
where $n$ is the number of agents and $\mathcal{E}$ is the set of edges of the coordination graph.
Note that without any assumptions on the interaction between the agents, solving this problem would be intractable as it would require searching the whole action space.
Instead, this problem can be solved approximately using the max-plus (or message passing) algorithm~\cite{kok2005maxplus}.
Each agent $i$ computes a message, $\mu_{ij} \in \mathbb{R}^{|\mathcal{A}_j|}$, for each of its neighbors.
This message depends on the value function of the edges connected to agent $i$ and on the messages received from its neighbors.It is computed as follows:
\begin{equation}
	\mu_{ij}(a_j) = \max_{a_i} [ Q_{ij}(s_i, s_j, a_i, a_j) + \sum_{k\in \mathcal{N}(i), k \neq j}\mu_{ki}(a_i)] + c_{ij}
	\label{eq:mp}
\end{equation}
for all $a_j \in \mathcal{A}_j$, where $\mathcal{N}(j)$ is the set of neighbors of $j$, and $c_{ij}$ is a normalizing term.
$Q_{ij}$ is parameterized by a neural network and is a function of the states.
However, during the message passing procedure, the state does not change. Hence, $Q_{ij}$ can be treated as a 2D matrix of size $|\mathcal{A}_i|\times|\mathcal{A}_j|$.
The agents keep sending, receiving, and recalculating messages until the value of the messages converges.
After convergence, individual actions are given by:
\begin{equation}
	a_i^* = \argmax_{a_i}\sum_{j\in\mathcal{N}(i)} \mu_{ji}(a_i)
\end{equation}
Exploration techniques like $\epsilon$-greedy can be applied for each agent.


\textbf{Learning.} The second step of the algorithm consists of learning the value functions at the edges of the graph.
Those value functions represent the pairwise influence of agents among each other.
Two subtleties occur in deriving an update rule for the value function: the credit assignment and the best action.
In standard RL we could use $r_i + r_j$ as a reward signal.
However, agent $i$ is connected to agent $j$ but also to other neighbors.
Hence, the performance of agent $i$ must be distributed among all of its neighbors as they might all contribute to its value (whether it is a good or bad influence).
The reward used to learn $Q_{ij}$ is then given by $\frac{r_i}{\mathcal{N}(i)} + \frac{r_j}{\mathcal{N}(j)}$.

The second aspect to consider is how to evaluate the target value.
If we use $\argmax_{a_i, a_j} Q_{ij}(s'_i, s'_j, ., .)$, the value function would ignore that the action selection procedure is influenced by the whole graph and not only the pair of interest.
To correct this effect, the target value is given by $Q_{ij}(s'_i, s'_j, a^*_i, a^*_j)$ where $a^*_i$, and $a^*_j$ are the results from the message passing algorithm described in the previous section.
We can now derive a generic update rule for coordinated RL with function approximation applied for all $i$, $j$:
\begin{equation}
	\theta \leftarrow \theta + \alpha[\frac{r_i}{\mathcal{N}(i)} + \frac{r_j}{\mathcal{N}(j)} + \gamma Q_{ij}(s'_i, s'_j, a^*_i, a^*_j; \theta)
	- Q_{ij}(s_i, s_j, a_i, a_j; \theta)]\nabla_{\theta}Q_{ij}(s_i, s_j, a_i, a_j;\theta)
	\label{eq:learning}
\end{equation}
where $\alpha$ is the learning rate.
This update rule can be carried with only local information about agents $i$ and $j$.
Similarly as in DQN, the parameters can be updated using batches of experience samples, and we can use additional improvements like double Q learning and target networks~\cite{hasselt2016deep}.
Relying on the discussion above, we could generalize the update rule to online learning methods which are often more sample efficient~\cite{bargiacchi2018learning} and would be more suitable for training in the real world.
A discussion and theoretical justification of the update rule in the tabular case is provided in previous work~\cite{kok2006collaborative}.

\section{Problem Formulation}

\begin{figure}
	\centering
	\includegraphics[width=\columnwidth]{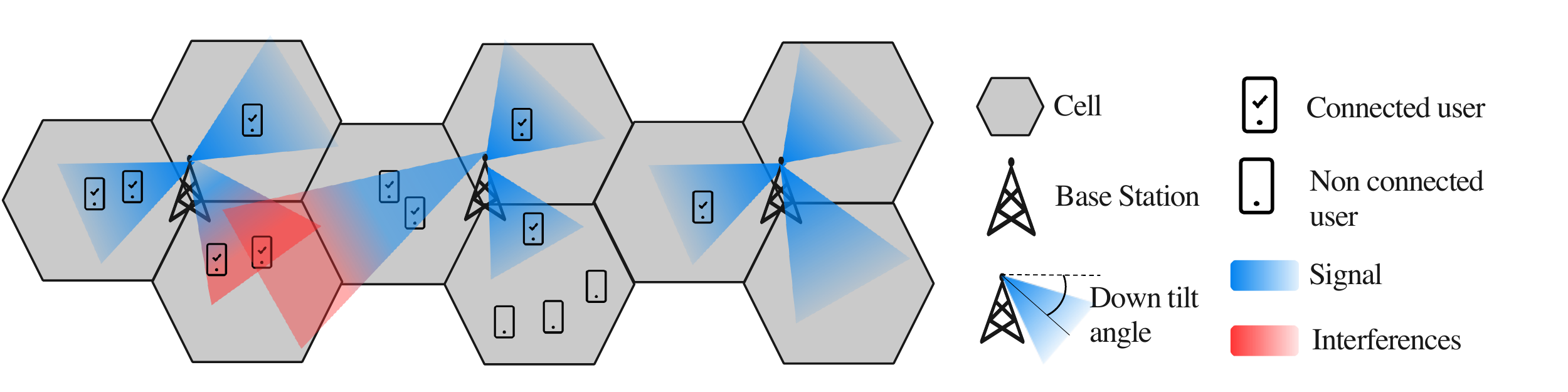}
	\caption{Illustration of a network optimization problem where the variable is antenna tilt. Each base station consists of three cells. We can see two undesirable events caused by poor configuration and lack of coordination.
	To the left, two antennas interfere with each other because of low tilt values in the middle base station (causing the signal to beam upward).
	In the bottom right, a high tilt value causes users to lose coverage.}
	\label{fig:cells-diagram}
\end{figure}

The problem of optimizing network performance can be modeled as a multi-agent MDP.
We consider a mobile telecommunication network consisting of multiple base stations each of them with multiple cells.
Each cell is associated to an antenna as illustrated in \cref{fig:cells-diagram}.
User equipments (users) are associated to a cell if they receive a sufficiently strong signal from the antenna of that cell.
The signal strength is affected by tunable parameters such as the tilt or the power of the antennas along with environmental aspects and ultimately affects the quality of experience of users.
In this work, we focus on downlink performance only and it is assumed that all users are active all the time.

In this paper, we choose to optimize antenna tilt configurations. However, our approach is applicable to any network parameter optimization.
Each base station consist of three cells with directional antennas.
Nevertheless, our approach can generalize to other types of deployments with different number of cells per base station and even heterogeneous base stations.
Each cell is associated to a reinforcement learning agent.

\Cref{fig:cells-diagram} illustrates one possible interaction between agents with the appearance of interference in the red area.
Two nearby antennas have low down tilt causing interference between their signals.
The agents must coordinate to alleviate the situation.
On the other hand, the antenna in the middle facing down has very high down tilt, causing a lack of coverage for users at the extremity of the cell.
Learning control policies for these agents requires coordination and collaboration to maximize the overall network quality for all users.
The multi-agent MDP is further specified as follows:

\textbf{Observed state.} Each agent observes the signal to interference and noise ratio (SINR), in \SI{}{\deci\bel}, of all the users it covers.
Since the number of users in a cell can vary over time, we transform the SINR list to a fixed vector representation by extracting four percentiles of the distribution: \SI{10}{\percent}, \SI{25}{\percent}, \SI{50}{\percent}, \SI{75}{\percent}.
We found that this representation captured well different characteristics of the cell coverage without making the observation space too complex. The observation space is then a four dimensional vector of real values.
The SINR is directly influenced by a change in antenna tilt and captures interference between agents.
Further details on how SINR is computed and how antenna gain is modeled can be found in appendix.

\textbf{Action space.}
We consider that the antenna in each cell can set its electrical down tilt, $\alpha$, to one of sixteen possible values in the set $\{\SI{0}{\degree}, \SI{1}{\degree}, \ldots, \SI{15}{\degree}\}$.
The action space of one agent is then a discrete space with sixteen possible actions.

\textbf{Reward function.}
The goal of our algorithm is to maximize the downlink throughput (in megabits per second) for every UE. Each cell can measure its contribution to this global objective by measuring the average throughput in this cell. We thus have one local reward signal per agent. Mathematically the overall objective is expressed as follows: $R(\mathbf{s}, \mathbf{a}) = \sum_{i=1}^n T_i$. Where $T_i$ is the logarithm of the average throughput in cell $i$.
The throughput is affected by the signal strength and the load of a given cell, if too many users are connected to a cell the throughput drops as they have to share resources.
Maximizing throughput globally in the network requires some coordination between all the agents.

Further details on the modeling is provided in appendix, including information about antenna models and formulas used to compute SINR and throughput. Although we only demonstrate the performance of the algorithm on the problem of antenna tilt control, it can generalize to other problems and supports heterogeneous deployments where each agent has a different action space.

\section{Mobile Networks as Coordination Graphs}

To take full advantage of the topology of the cellular networks we propose to represent it using a coordination graph.
Given a deployment of base stations, we construct a graph where each cell is a node.
An edge is assigned between cells that can influence the received signal of each other's users.
To determine such relations we can use automated procedure using domain knowledge and heuristics: based on the geographic distance between cells, using the radiation patterns of the antennas~\cite{saeed2012controlling}, using automatic neighbor relations (ANR) as defined in international standards~\cite{aliu2013survey}, or using network planning tools computing coverage prediction.

\begin{figure}
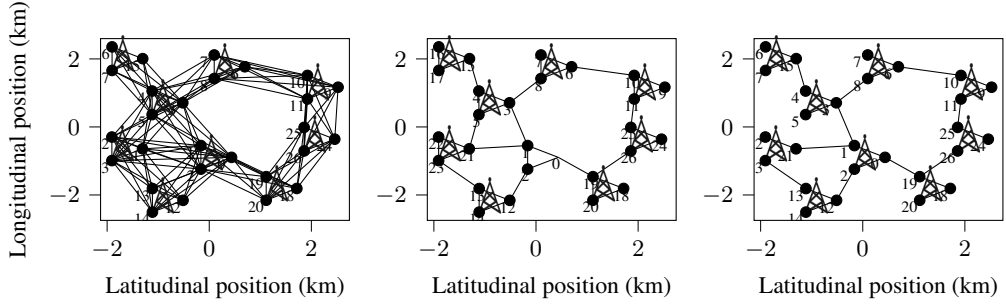

	\centering
	\input{figures/dense_coordination_graph.tex}
  \input{figures/coordination_graph.tex}
  \begin{tikzpicture}

    \definecolor{color0}{rgb}{0,0.509803921568627,0.941176470588235}

    \begin{axis}[
                height=4cm,
        width=0.35\columnwidth,
    tick align=outside,
    tick pos=left,
    x grid style={white!69.0196078431373!black},
    xlabel={Latitudinal position (km)},
    xmin=-2.12832935379154, xmax=2.74482572023113,
    xtick style={color=black},
    y grid style={white!69.0196078431373!black},
    ymin=-2.74722604329433, ymax=2.59691193189433,
    ytick style={color=black}
    ]
    \path [draw=black]
(axis cs:0.436505186271202,-0.894083422938598)
--(axis cs:-0.163494813728798,-0.547673261424823);

  \path [draw=black]
  (axis cs:0.436505186271202,-0.894083422938598)
  --(axis cs:-0.163494813728798,-1.24049358445237);

  \path [draw=black]
  (axis cs:0.436505186271202,-0.894083422938598)
  --(axis cs:1.11545302375113,-1.46225486993517);

  \path [draw=black]
  (axis cs:-0.163494813728798,-0.547673261424823)
  --(axis cs:-0.518266228710206,0.709268964654527);

  \path [draw=black]
  (axis cs:-0.163494813728798,-0.547673261424823)
  --(axis cs:-1.30682230497233,-0.643368238248841);

  \path [draw=black]
  (axis cs:-0.163494813728798,-1.24049358445237)
  --(axis cs:-0.512906445214897,-2.15790051927198);

  \path [draw=black]
  (axis cs:-0.518266228710206,0.709268964654527)
  --(axis cs:-1.11826622871021,1.0556791261683);

  \path [draw=black]
  (axis cs:-0.518266228710206,0.709268964654527)
  --(axis cs:-1.11826622871021,0.362858803140752);

  \path [draw=black]
  (axis cs:-0.518266228710206,0.709268964654527)
  --(axis cs:0.0974386682868965,1.42285589483522);

  \path [draw=black]
  (axis cs:-1.11826622871021,1.0556791261683)
  --(axis cs:-1.30296981602234,2.00758640787198);

  \path [draw=black]
  (axis cs:0.697438668286897,1.76926605634899)
  --(axis cs:0.0974386682868968,2.11567621786277);

  \path [draw=black]
  (axis cs:0.697438668286897,1.76926605634899)
  --(axis cs:0.0974386682868965,1.42285589483522);

  \path [draw=black]
  (axis cs:0.697438668286897,1.76926605634899)
  --(axis cs:1.92331867141192,1.51428266029856);

  \path [draw=black]
  (axis cs:2.52331867141192,1.16787249878479)
  --(axis cs:1.92331867141192,1.51428266029856);

  \path [draw=black]
  (axis cs:2.52331867141192,1.16787249878479)
  --(axis cs:1.92331867141192,0.821462337271011);

  \path [draw=black]
  (axis cs:1.92331867141192,0.821462337271011)
  --(axis cs:1.85808222154808,-0.0133193013196898);

  \path [draw=black]
  (axis cs:-0.512906445214897,-2.15790051927198)
  --(axis cs:-1.1129064452149,-1.8114903577582);

  \path [draw=black]
  (axis cs:-0.512906445214897,-2.15790051927198)
  --(axis cs:-1.1129064452149,-2.50431068078575);

  \path [draw=black]
  (axis cs:-1.1129064452149,-1.8114903577582)
  --(axis cs:-1.90682230497233,-0.989778399762616);

  \path [draw=black]
  (axis cs:-1.30296981602234,2.00758640787198)
  --(axis cs:-1.90296981602234,2.35399656938576);

  \path [draw=black]
  (axis cs:-1.30296981602234,2.00758640787198)
  --(axis cs:-1.90296981602234,1.66117624635821);

  \path [draw=black]
  (axis cs:1.71545302375113,-1.80866503144894)
  --(axis cs:1.11545302375113,-1.46225486993517);

  \path [draw=black]
  (axis cs:1.71545302375113,-1.80866503144894)
  --(axis cs:1.11545302375113,-2.15507519296272);

  \path [draw=black]
  (axis cs:1.11545302375113,-1.46225486993517)
  --(axis cs:1.85808222154808,-0.706139624347241);

  \path [draw=black]
  (axis cs:-1.30682230497233,-0.643368238248841)
  --(axis cs:-1.90682230497233,-0.296958076735065);

  \path [draw=black]
  (axis cs:2.45808222154808,-0.359729462833465)
  --(axis cs:1.85808222154808,-0.0133193013196898);

    \addplot [only marks, mark=bs, draw=color0, fill=color0, colormap/viridis]
    table{%
    x                      y
    0.0365051862712021 -0.744083422938598
    -0.918266228710206 0.859268964654527
    0.297438668286897 1.91926605634899
    2.12331867141192 1.31787249878479
    -0.912906445214897 -2.00790051927198
    -1.70296981602234 2.15758640787198
    1.31545302375113 -1.65866503144894
    -1.70682230497233 -0.493368238248841
    2.05808222154808 -0.209729462833465
    };
    \addplot [only marks, mark=*, draw=black, fill=black, colormap/viridis]
    table{%
    x                      y
    0.436505186271202 -0.894083422938598
    -0.163494813728798 -0.547673261424823
    -0.163494813728798 -1.24049358445237
    -0.518266228710206 0.709268964654527
    -1.11826622871021 1.0556791261683
    -1.11826622871021 0.362858803140752
    0.697438668286897 1.76926605634899
    0.0974386682868968 2.11567621786277
    0.0974386682868965 1.42285589483522
    2.52331867141192 1.16787249878479
    1.92331867141192 1.51428266029856
    1.92331867141192 0.821462337271011
    -0.512906445214897 -2.15790051927198
    -1.1129064452149 -1.8114903577582
    -1.1129064452149 -2.50431068078575
    -1.30296981602234 2.00758640787198
    -1.90296981602234 2.35399656938576
    -1.90296981602234 1.66117624635821
    1.71545302375113 -1.80866503144894
    1.11545302375113 -1.46225486993517
    1.11545302375113 -2.15507519296272
    -1.30682230497233 -0.643368238248841
    -1.90682230497233 -0.296958076735065
    -1.90682230497233 -0.989778399762616
    2.45808222154808 -0.359729462833465
    1.85808222154808 -0.0133193013196898
    1.85808222154808 -0.706139624347241
    };
    \draw (axis cs:0.236505186271202,-1.0940834229386) node[
      scale=0.6,
      text=black,
      rotate=0.0
    ]{0};
    \draw (axis cs:-0.363494813728798,-0.747673261424823) node[
      scale=0.6,
      text=black,
      rotate=0.0
    ]{1};
    \draw (axis cs:-0.363494813728798,-1.44049358445237) node[
      scale=0.6,
      text=black,
      rotate=0.0
    ]{2};
    \draw (axis cs:-0.718266228710206,0.509268964654527) node[
      scale=0.6,
      text=black,
      rotate=0.0
    ]{3};
    \draw (axis cs:-1.31826622871021,0.855679126168303) node[
      scale=0.6,
      text=black,
      rotate=0.0
    ]{4};
    \draw (axis cs:-1.31826622871021,0.162858803140752) node[
      scale=0.6,
      text=black,
      rotate=0.0
    ]{5};
    \draw (axis cs:0.497438668286897,1.56926605634899) node[
      scale=0.6,
      text=black,
      rotate=0.0
    ]{6};
    \draw (axis cs:-0.102561331713103,1.91567621786277) node[
      scale=0.6,
      text=black,
      rotate=0.0
    ]{7};
    \draw (axis cs:-0.102561331713103,1.22285589483522) node[
      scale=0.6,
      text=black,
      rotate=0.0
    ]{8};
    \draw (axis cs:2.32331867141192,0.967872498784787) node[
      scale=0.6,
      text=black,
      rotate=0.0
    ]{9};
    \draw (axis cs:1.72331867141192,1.31428266029856) node[
      scale=0.6,
      text=black,
      rotate=0.0
    ]{10};
    \draw (axis cs:1.72331867141192,0.621462337271011) node[
      scale=0.6,
      text=black,
      rotate=0.0
    ]{11};
    \draw (axis cs:-0.712906445214897,-2.35790051927198) node[
      scale=0.6,
      text=black,
      rotate=0.0
    ]{12};
    \draw (axis cs:-1.3129064452149,-2.0114903577582) node[
      scale=0.6,
      text=black,
      rotate=0.0
    ]{13};
    \draw (axis cs:-1.3129064452149,-2.70431068078575) node[
      scale=0.6,
      text=black,
      rotate=0.0
    ]{14};
    \draw (axis cs:-1.50296981602234,1.80758640787198) node[
      scale=0.6,
      text=black,
      rotate=0.0
    ]{15};
    \draw (axis cs:-2.10296981602234,2.15399656938576) node[
      scale=0.6,
      text=black,
      rotate=0.0
    ]{16};
    \draw (axis cs:-2.10296981602234,1.46117624635821) node[
      scale=0.6,
      text=black,
      rotate=0.0
    ]{17};
    \draw (axis cs:1.51545302375113,-2.00866503144894) node[
      scale=0.6,
      text=black,
      rotate=0.0
    ]{18};
    \draw (axis cs:0.915453023751128,-1.66225486993517) node[
      scale=0.6,
      text=black,
      rotate=0.0
    ]{19};
    \draw (axis cs:0.915453023751128,-2.35507519296272) node[
      scale=0.6,
      text=black,
      rotate=0.0
    ]{20};
    \draw (axis cs:-1.50682230497233,-0.843368238248841) node[
      scale=0.6,
      text=black,
      rotate=0.0
    ]{21};
    \draw (axis cs:-2.10682230497233,-0.496958076735065) node[
      scale=0.6,
      text=black,
      rotate=0.0
    ]{22};
    \draw (axis cs:-2.10682230497233,-1.18977839976262) node[
      scale=0.6,
      text=black,
      rotate=0.0
    ]{23};
    \draw (axis cs:2.25808222154808,-0.559729462833465) node[
      scale=0.6,
      text=black,
      rotate=0.0
    ]{24};
    \draw (axis cs:1.65808222154808,-0.21331930131969) node[
      scale=0.6,
      text=black,
      rotate=0.0
    ]{25};
    \draw (axis cs:1.65808222154808,-0.906139624347241) node[
      scale=0.6,
      text=black,
      rotate=0.0
    ]{26};
    \end{axis}

    \end{tikzpicture}
	\caption{Example of a coordination graph constructed using interference patterns (middle). The position of the nodes is offset for the clarity of the visualization. The nodes close to a base station corresponds to antennas located at that base station. On the left, we consider a dense graph construction scheme. On the right we consider the minimum spanning tree of the graph.}
	\label{fig:cg}
\end{figure}

All these methods could be used to automate the construction of an undirected graph from a given deployment. In addition, domain knowledge can be used to refine the graph topology by pruning or adding edges based on key feature of a city or knowledge about the terrain (if there is a natural obstacle for example).
Reducing the number of edges can speed up the message passing procedure described in the previous section.
In this paper we construct the graph based on the radiation pattern of the antennas, that is antennas that can interfere with each other are connected by an edge.
An example of such graph is illustrated on a 27 cells deployment in \cref{fig:cg} (middle).

In case where the graph construction leads to a disconnected graph, then our algorithm can run in each connected component independently.
A disconnected graph would mean that there is no a priori interaction between its connected components.
Without loss of generality we assume a connected graph for the remaining of the article.
In \cref{sec:results} we propose a comparison with other graph construction techniques leading to a dense graph or a tree as illustrated in \cref{fig:cg}.
We found that adding edges did not yield any gain in performance.

Given a network deployment we can extract a graph in a simple way, yet capturing a lot of information about the interaction between agents.
Finding the optimal graph representation of a telecommunication network is an interesting problem and is left as future work.
As explained in the previous section, the graph representation can directly be used in a multi-agent reinforcement learning algorithm to perform network optimization at scale.

\textbf{Message Passing.} In our experiments, the graphs contain cycles and hence the message passing algorithm is only proven to converge empirically.
We implemented two stopping criteria for the algorithm, leading to an approximately optimal solution.
The first criterion is to stop the algorithm after a limited number of iterations.
Our second criterion detects the occurrence of oscillation in the values of the message, and compare the global value of each solution in the oscillation cycle.
The algorithm stops and assigns the action found by the best possible configuration from the detected cycle.
This early stopping criteria allows speeding up the message passing algorithm while making sure we stop at the best possible configuration found throughout the message exchange.
Future work would be needed in studying the performance gap between solving \cref{eq:inference} exactly or approximately.
We show in our experiment that even with the approximated method, the overall coordinated RL procedure is able to outperform many baselines.
An example of what messages look like in the antenna tilt use case is given in \cref{fig:message-passing}.

\textbf{Implementation considerations.}
In this paper we rely on a simulated environment and on a fully centralized implementation of the message passing algorithm.
However, an implementation in a real network could use the property that the message passing algorithm can be implemented in a fully distributed and anytime way~\cite{kok2005maxplus}.
The messages could be sent through existing interfaces for communicating between the base stations.
Since they only rely on local information, the RL updates could be carried in hardware located at the station.
In such a setting, the data would stay local to the network, alleviating privacy concern related to sending user data to a central server.

\section{Experiments}\label{sec:results}

We empirically demonstrate the advantage of coordinated RL over state-of-the art multi-agent RL algorithms and against heuristics in a high fidelity mobile network simulator. By comparing the performance in networks of various sizes we validate the scalability of the approach. Then we analyze the effect of the coordination graph topology on the final performance. Finally, we show that the learned edge value functions match intuitive coordination behavior between neighboring antennas.

\textbf{Simulation Environment.} We evaluate our approach using a high fidelity proprietary network simulator that handles multiple cells and multiple antennas. The simulator performs all the path gain calculations using advanced radio propagation models as well as traffic calculations for each user. From that simulator we can get different key performance indicators for each user such as SINR and throughput. Using the simulator, we generate different deployments from \num{6} to \num{207} cells.
The base stations are positioned according to a Poisson process with a minimum intersite distance of \SI{1.5}{\kilo\meter}.
The simulated area increases as well such that the average intersite distance stays the same. More details on the environment are provided in appendix.

\textbf{Baselines.} Our approach is referred to as PS-CRL for coordinated RL with parameter sharing. We compare it to six different baselines:
\begin{itemize}
  \item Coordinated RL (CRL): our approach without parameter sharing. A different neural network is used for each edge of the graph.
  \item Independent DQN (I-DQN): this algorithm assumes that all cells are independent and associate a Q-network for each of them~\cite{tan1993multi}. Each Q-network is learned independently.
  \item DQN with parameter sharing (S-DQN) this algorithm associates the same Q-network for each cell. It is similar to I-DQN but the weights of the network are the same for each cell. It has been shown to greatly help in problem with homogenous cooperating agents~\cite{gupta2017cooperative}.
  \item QMIX: this algorithm attempts to capture agent coordination by learning a non-linear combination of individual value functions, modeled by a neural network~\cite{rashid2018qmix}.
  \item sweep: This algorithm is not a learning based method. For each cell, we sweep through all the tilt values while maintaining other cells fixed, and choose the best tilt value. It is agnostic of the state of the environment.
  \item C-sweep: We perform a tilt sweep for each pair of connected agent according to the same graph topology as in CRL. When doing the sweeping we populate a table of size $|\mathcal{A}_i|\times|\mathcal{A}_j|$ for each edge. Message passing is then used to choose the best action.
\end{itemize}

\subsection{Training Performance and Scalability}
\label{sec:training-performance}\label{sec:scalability}

\begin{figure}
	\centering
	\input{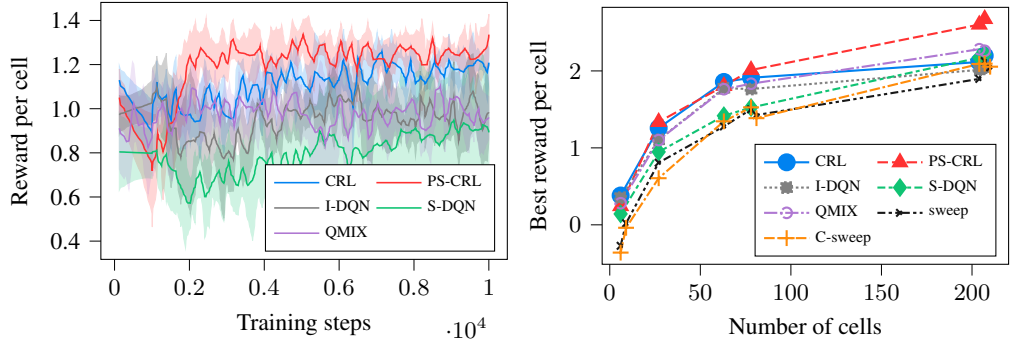}
	\begin{tikzpicture}

    \definecolor{color0}{rgb}{0,0.509803921568627,0.941176470588235}
    \definecolor{color1}{rgb}{1,0.196078431372549,0.196078431372549}
    \definecolor{color2}{rgb}{0.980392156862745,0.823529411764706,0.176470588235294}
    \definecolor{color3}{rgb}{0.0588235294117647,0.764705882352941,0.450980392156863}
    \definecolor{color4}{rgb}{0.686274509803922,0.470588235294118,0.823529411764706}
    \definecolor{color5}{rgb}{1,0.549019607843137,0.0392156862745098}

    \begin{axis}[
        height=5cm,
        width=0.5\columnwidth,
        legend columns=2,
    legend cell align={left},
    legend style={at={(0.97,0.03)}, anchor=south east, font=\tiny},
    legend entries={CRL, PS-CRL, I-DQN, S-DQN, QMIX, sweep, C-sweep},
    tick align=outside,
    tick pos=left,
    x grid style={white!69.0196078431373!black},
    xlabel={Number of cells},
    xmin=-4.2, xmax=220.2,
    xtick style={color=black},
    y grid style={white!69.0196078431373!black},
    ylabel={Best reward per cell},
    ymin=-0.512585790568045, ymax=2.8261124511281,
    ytick style={color=black}
    ]
    \addplot [thick, color0, mark=*, mark size=3, mark options={solid,draw=color0}]
    table {%
    6 0.381098000208537
    27 1.25812126442238
    63 1.85684468102834
    78 1.91073028564453
    204 2.11636202482796
    207 2.19744906218156
    };
    \addplot [thick, color1, dash pattern=on 4pt off 1.5pt, mark=triangle*, mark size=3, mark options={solid,draw=color1}]
    table {%
    6 0.243057894706726
    27 1.3366086324056
    63 1.80452549041264
    78 2.01155473758013
    204 2.60209945977903
    207 2.67435344014191
    };
    \addplot [thick, gray, dash pattern=on 1pt off 1pt, mark=square*]
    table {%
    6 0.352470239003499
    27 1.10490671408297
    63 1.77517692929222
    78 1.76460888984876
    204 2.01362741507736
    207 2.04299171926895
    };
    \addplot [thick, color3, dash pattern=on 3pt off 1.25pt on 1.5pt off 1.25pt, mark=diamond*, mark size=3, mark options={solid,draw=color3}]
    table {%
    6 0.142021920283635
    27 0.947494836501133
    63 1.41740407792349
    78 1.52431943844526
    204 2.17216088454968
    207 2.217338082871
    };
    \addplot [thick, color4, dash pattern=on 5pt off 1pt on 1pt off 1pt, mark=o]
    table {%
    6 0.280754554271698
    27 1.10102685292562
    63 1.7622564648825
    78 1.8337153508113
    204 2.28203920850567
    207 2.26154847075974
    };
    \addplot [thick, white!9.41176470588235!black, dash pattern=on 3pt off 1.25pt on 1.25pt off 1.25pt on 1.25pt off 1.25pt, mark=x]
    table {%
    6 -0.27266458355171
    9 0.0206077020744095
    27 0.807824205542528
    63 1.27912513244964
    78 1.50057073421267
    81 1.42659676413406
    204 1.89643992316359
    207 2.12730089180971
    210 2.04690025919598
    };
    \addplot [thick, color5, dash pattern=on 4pt off 1pt on 4pt off 1pt on 1pt off 1pt, mark=+, mark size=3, mark options={solid,draw=color5}]
    table {%
    6 -0.360826779581856
    9 -0.038379854999287
    27 0.605875865911058
    63 1.3447819038557
    78 1.5264531773832
    81 1.38613633085233
    204 2.09058367698038
    207 2.09431978911561
    210 2.05493036251049
    };
    \end{axis}

    \end{tikzpicture}
	\caption{(left) Training performance of the RL methods on a network with \num{27} cells. (right) Reward per cells as a function of the total number of cells for the best trained models.}
	\label{fig:reward-per-nbs}
\end{figure}

We measure the performance of the trained model in terms of average reward (evaluated across \num{10} different user distributions for every point) which is directly related to the average throughput. \Cref{fig:reward-per-nbs} left illustrates the training performance of our method and the baseline. Each experiment is repeated with five different random seeds. During training, we use $\epsilon$-greedy exploration but we report the evaluated reward (without $\epsilon$-greedy) at different checkpoints. We show that on this network with \num{27} agents, both PS-CRL and CRL outperforms all the baselines in terms of final reward, convergence speed and show less variance.

\Cref{fig:reward-per-nbs} shows the performance of the best performing models for different network sizes.
We choose the models with the best average reward throughout the training process (not necessarily the latest).
We are reporting the average reward per cell as a function of the number of cells in the environment. One cell corresponds to one RL agent.
We notice that there is a monotonic trend, as the number of cell increases, the reward per cell seems to increase.
This matches our intuition since more cells should imply better coverage.
However, adding cells can also create interference, hence the small degradation for I-DQN, sweep, and C-sweep around 78 cells.
We can see that all the way to 78 cells, CRL and PS-CRL are outperforming all the baselines.
As the number of cell increases we observe a similar fact than in the previous section.
PS-CRL reaches the best performance, and CRL is getting caught up by S-DQN and QMIX.
In addition, it seems like the benefit of using coordination increases with the number of cells as the gap between the performance of PS-CRL and the other method is increasing as well.

Finally, we introduced two non-learning based method in this plot, sweep and C-sweep.
They are generally outperformed by the RL baseline, until the environment becomes large enough that the convergence of the RL policy is affected.
These empirical results show the benefit of CRL and PS-CRL in order to model coordination between cells and train multi-agent RL algorithms at scale on telecommunication networks.
Our largest deployments have an order of magnitude more agents than in standard multi-agent RL algorithms benchmarks.

\subsection{Interpreting Edge Value Functions}

\begin{figure}[t]
	\centering
	\input{figures/graph_structure_training.tex}
	\includegraphics[width=0.45\columnwidth]{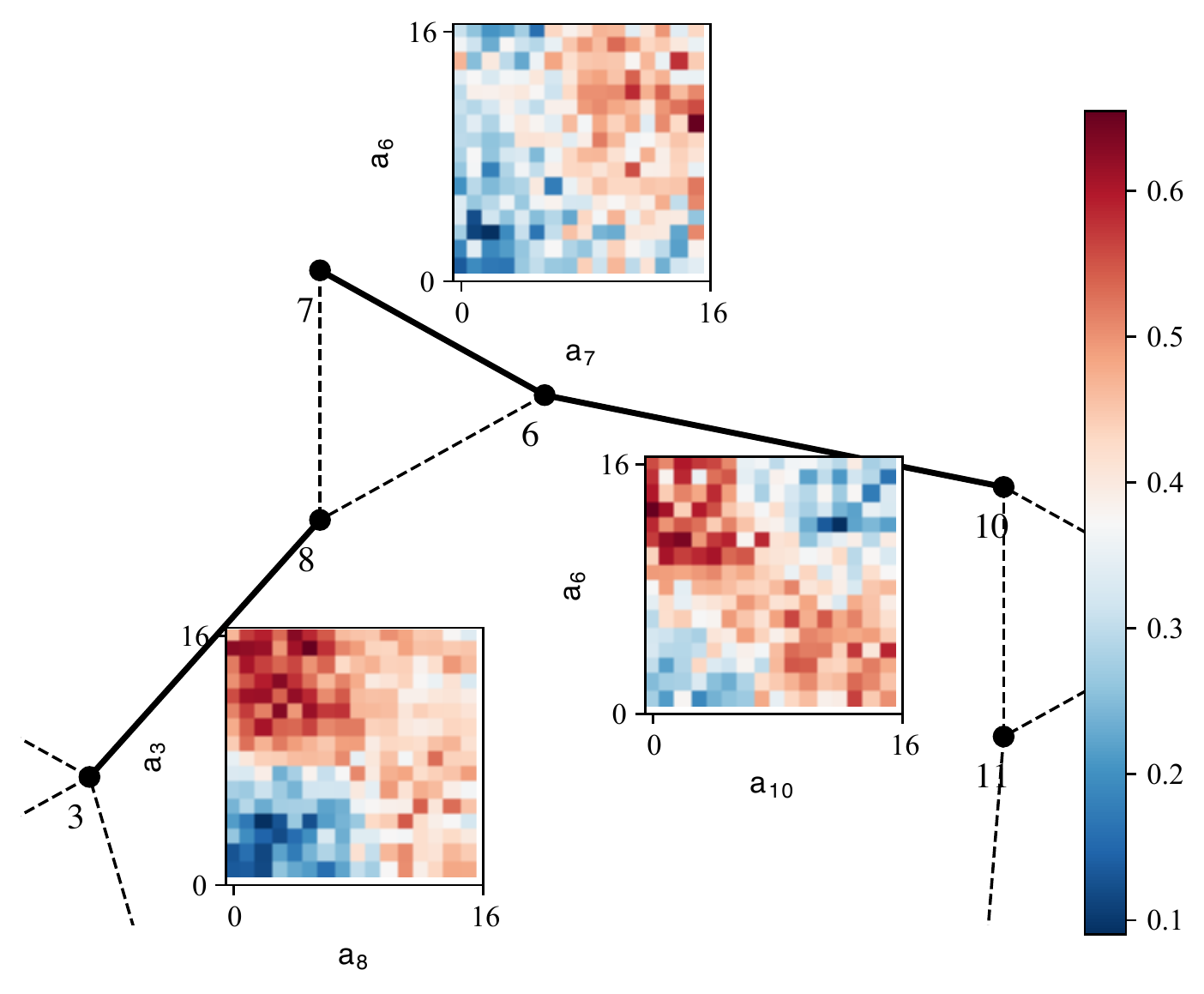}
	\caption{(left) Performance of PS-CRL for different types of coordination graphs on a scenario with \num{78} agents. (right) Illustration of the learned value functions for the three solid edges in a subset of the sparse graph from \cref{fig:cg}.}
	\label{fig:value-functions}
\end{figure}

\Cref{fig:value-functions} illustrates the value functions for certain edges in a given state.
Each axis on the heatmap corresponds to the action of one of the two connected agents.
The bottom left value function (edge (3,8)), shows that there is low value when both agents have low tilt (bottom left corner).
In that case, the antennas are likely to interfere with each other.
The right most value function (edge (6,10)) also shows that when both antennas have high tilt (top right corner), the value is low.
In that case, antennas are tilted down, there is no interference, but the coverage is poor.
The learned value functions align with the intuition that for two neighboring antennas both should not be tilted up or down at the same time. The ability to interpret the value function by visualizing them as a heatmap gives great value to the coordination graph representation.
Experts can look at the learned value function across the graph and diagnose which antennas have strong correlations and use it as a network planning tool.

\subsection{Graph Structure}

Different methods for creating a coordination graphs are considered: sparse, dense and tree as illustrated in \cref{fig:cg}, and a fully connected graph (complete).
We analyze the performance of PS-CRL for these different graph structures.
We can see that for the dense, sparse, and tree methods, PS-CRL is converging to similar values. When there are too many connections, it fails to learn a good solution.
In problems where graph structure can be identified PS-CRL is a good solution even if the graph model is not accurate.
In the tree representation, information can still flow between distant agents through the message passing procedure even if they are not directly connected.

\section{Conclusions}

Coordinated RL with parameter sharing allows learning control strategies for a large number of cooperating agents.
Its ability to exploit graph structures makes it a good candidate for mobile network optimization problem.
We presented a principled way to model mobile networks as coordination graphs and learn a value function for each edge of this graph.
By leveraging the topology of the graph, an efficient message passing algorithm can be used to coordinate all the learning agents.
We showed that the proposed algorithm outperforms a variety of RL and non RL baselines.
The learned value functions at the edge of the graph can be interpreted and were shown to capture intuitive phenomena about the network.
When augmented with parameter sharing, CRL has better sample efficiency and scales to hundreds of learning agents.

One of the main challenges of applying RL to telecommunication networks is that it requires a lot of data.
Even though simulators are available, they are often slow.
Future work involves looking at speeding up simulation and mixing simulated and real world data.
Another interesting direction involves transferring learned policies from one graph to another.


\printbibliography


\appendix

\section{Additional Details on the Simulation Environment}

The base stations we simulate are LTE base stations, operating at \SI{2}{\giga\hertz}. Antennas are at a \SI{32}{\meter} height, have a maximum transmitting power of \SI{40}{\watt}, a horizontal beamwidth of \SI{65}{\degree} and a vertical beamwidth of \SI{6.5}{\degree}.
The radiation pattern of the antenna is fitted from real base station antenna data.

Using the simulator, we generate different deployments from \num{6} to \num{207} cells.
The base stations are positioned according to a Poisson process with a minimum intersite distance of \SI{1.5}{\kilo\meter}.
The simulated area increases as well such that the average intersite distance stays the same.
The environment is generated randomly such that \SI{50}{\percent} of the area corresponds to indoor environments and the rest outdoor.
There are \num{1000} users uniformly distributed in the environment and their position is re-sampled after every change in tilt configuration.

\subsubsection*{Observed state}
Each agent observes the signal to interference and noise ratio (SINR), in \SI{}{\deci\bel}, of all the users it covers.
Since the number of users in a cell can vary over time, we transform the SINR list to a fixed vector representation by extracting four percentiles of the distribution: \SI{10}{\percent}, \SI{25}{\percent}, \SI{50}{\percent}, \SI{75}{\percent}.
We found that this representation captured well different characteristics of the cell coverage without making the observation space too complex.
The observation space is then a four dimensional vector of real values.
The choice of SINR is motivated by the fact that it is directly influenced by a change in antenna tilt and captures interference from other cells.
The downlink SINR, $\rho_{c,u}$, of a user $u$ connected to cell $c$ is expressed as the ratio of the received power measured by $u$ from cell $c$ and the sum of the received power from all other cells, and the noise power~$\sigma$~\cite{asghar2018concurrent}, assuming full load.
\begin{equation}
	\rho_{c,u} = \frac{P_{c}G_{c,u}L_{c,u}}{\sum_{i=1, i\neq c }^{N_{\text{cells}}} P_{i}G_{i,u}L_{i,u} + \sigma}
\end{equation}
$P_{c}$, $G_{c,u}$, $L_{c,u}$ are respectively the transmitted power, gains of the transmitter antenna, and path loss for UE $u$ connected to cell $c$.
The denominator accounts for the effect of other cells, the gain $G$ is influenced by antenna parameters such as tilt and azimuth, and the path loss accounts for the transmission medium and obstacles (building, air, trees, ...). We use a prioprietary antenna model which is fitted using real antenna data.

\subsubsection*{Action space} Each cell contains an antenna. The signal emission pattern of this antenna can be adjusted by a variety of parameters. Changing those parameters will increase or degrade the capacity and coverage of the cell.
In this paper we focus on electrical down tilt as it has one of the biggest effect on network key performance indicators and can easily be controlled by an automated system (contrary to mechanical tilt).
We consider that the antenna in each cell can set its electrical down tilt, $\alpha$, to one of sixteen possible values in the set $\{\SI{0}{\degree}, \SI{1}{\degree}, \ldots, \SI{15}{\degree}\}$.
The action space of one agent is then a discrete space with sixteen possible actions.

\subsubsection*{Reward} The goal of our algorithm is to maximize coverage and capacity. Our algorithm aims at maximizing throughput. We consider the throughput of a UE to be defined by the bitrate of that user divided by the load of the cell.
The bitrate is the maximum amount of data per second that a user can receive given the SINR produced by its associated antenna.
The load of the cell is equal to the number of users covered by a given cell.
Since the bitrate can take very large values spread across a wide range, we use the logarithm of the average bitrate in a cell in our reward definition.
The overall objective of our cooperative multi-agent RL problem is expressed mathematically as follows:
\begin{equation}
	R(\mathbf{s}, \mathbf{a}) = \sum_{i=1}^{N_{\text{cells}}} [\log(\frac{1}{|\mathbb{U}_i|}\sum_{u\in \mathbb{U}_i} T_{i,u})]
	\label{eq:reward}
\end{equation}
where $N_{\text{cells}}$ is the number of cells in the network, $T_{i,u}$ is the throughput of user $u$ connected to cell $i$, and $\mathbb{U}_i$ is the set of all users in cell $i$.
The throughput is expressed as a function of the SINR and bandwidth available to user $u$:
\begin{equation}
	T_{i,u} = n_{i,u}\omega_B f(\rho_{i,u})
	\label{eq:throughput}
\end{equation}
where $f(\rho_{i,u})$ is the spectral efficiency of the user link for a given SINR, $n_{i,u}$ is the number of physical resource blocks (PRBs) allocated to user $u$ in cell $i$ and $\omega_B$ is the bandwidth per PRB (\SI{180}{\kilo\hertz}).
Considering round robin scheduling (equal number of PRB allocated to each user) and the Shanon rate capacity, \cref{eq:throughput} becomes:
\begin{equation}
	T_{i,u} = \frac{n_B \omega_B}{|\mathbb{U}_i|}\log_2(1 + \rho_{i,u})
\end{equation}
where $n_B$ is the total number of PRBs in cell $i$.
The number of users connected to cell $i$ is also affected by tilt variation since we assume users connect to the cell from which they get maximum reference signal received power (RSRP).
Given a tilt value $\alpha$, the users in cell $i$ are defined by:
\begin{equation}
	U_i = \{ \forall u \in U \mid i = \argmax_{\forall c\in C} P_{c} G_{c,u}(\alpha) L_{c, u}\}
\end{equation}

\section{Additional Details on Training Methodology}

The observation and reward are the same for all the methods considered.
In order to have good numerical conditions we normalize the observation vector by dividing by the maximum SINR value.
The reward function is normalized to have zero mean and unit variance.
The mean and variance of the reward, along with the maximum SINR, are estimated through \num{1000} random tilt configuration.

All the baselines are part of the RLlib python package and our method is implemented using the same abstractions~\cite{liang2018rllib}.
This library allows us to distribute the experience collection across multiple CPUs.
We performed hyperparameter search on the learning rate, batch size, and target network update frequency and the exploration schedule for each method.
The network architecture was tuned using the I-DQN algorithm and we then use networks of equivalent sizes for all the algorithms.
All the hyperparameters are reported in \cref{tab:hyperparameters}, parameters that are not specified are set to the default provided by RLlib.
Each method is allowed a budget of \num{10000} interactions with the simulation environment and each training is repeated with \num{5} different random seeds.

For all baselines, the experience collection strategy is similar to the deep Q-learning algorithm with prioritized experience replay~\cite{mnih2015human, schaul2016prioritized}. For exploration, we rely on an $\epsilon$-greedy policy. In the case of shared DQN, we use the same replay buffer for all the agents.

\begin{table}
	\footnotesize
	\centering
	\caption{Hyperparameters of the deep RL algorithms}
	\label{tab:hyperparameters}
	\begin{tabular}{lll}
		\toprule
		\multirow{8}{*}{Common hyperparameters} \\
		{} & $\gamma$ & \num{0.0} \\
		{} & learning rate & \num{1e-4} \\
		{} & initial $\epsilon$ & 1.0 \\
		{} & final $\epsilon$ & 0.01 \\
		{} & $\epsilon$ decrease steps & \num{5000} \\
		{} & hidden layers & $32\times32$ \\
		{} & activation function & ReLU \\
		{} & batch size & 32 \\
		\midrule
		\multirow{1}{*}{DQN specific} \\
		{} & target network update frequency & \num{2000} \\
		\midrule
		\multirow{2}{*}{CRL specific} \\
		{} & message passing max iterations & 40 \\
		{} & target network update frequency & 500 \\
		\midrule
		\multirow{2}{*}{QMIX specific} \\
		{} & mixing embedding dimension & 32 \\
		{} & double Q learning & yes \\
		\bottomrule
	\end{tabular}
\end{table}

\section{Comparison of Inference Time for Different Graph Models}

\begin{figure}
	\centering
	\input{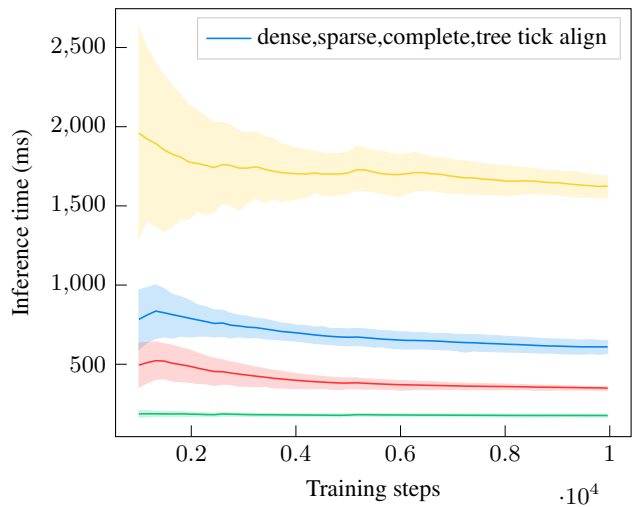}
	\label{fig:inference}
	\caption{Inference time for PS-CRL throughout training for different types of coordination graphs.}
\end{figure}

The inference time of CRL depends on the size of the graph in two ways. First once must compute the value function for each edge at a given state which is linear in the number of edges. Then the message passing procedure also depends on the graph size. One round of message passing is given by \cref{eq:mp}, which is linear in the degree of a node.
The more complex the graph, the more message passing iteration might be needed at each action selection step as well. A theoretical analysis is out of the scope of this work. In \cref{fig:inference} we illustrate the influence of the coordination graph on the inference time. As expected, the less edges the faster (for the same number of nodes). Since dense, sparse, and tree provides similar performance, a tree like graph would be the most efficient since it has a shorter inference time.

\section{Background on Message Passing}

\begin{figure}[t]
	\centering
	\includegraphics[width=0.8\columnwidth]{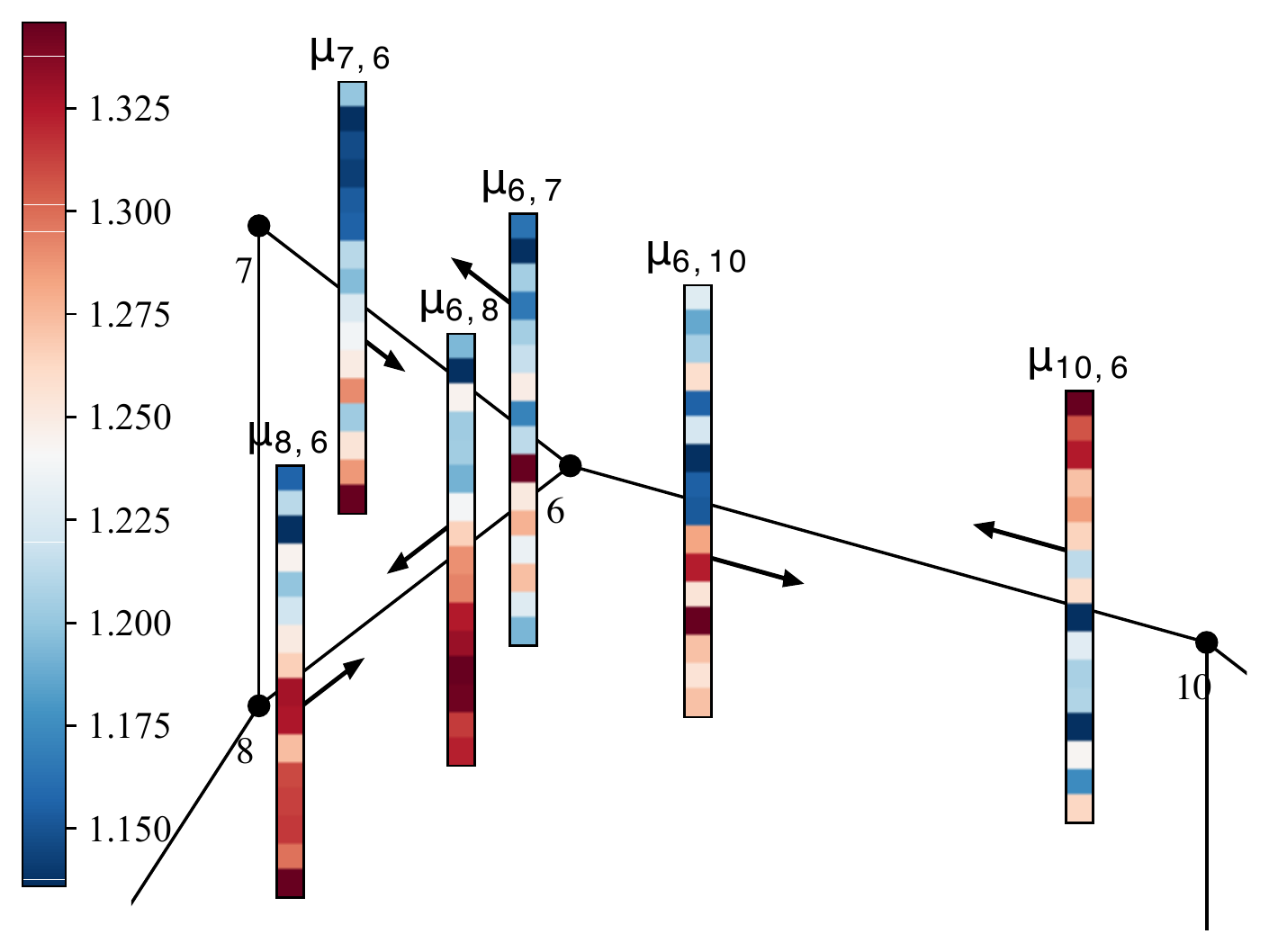}
	\label{fig:message-passing}
	\caption{Illustration of the message passing procedure from the point of view of node 6. The messages sent and received have the size of the action space of the receiving neighbor. In this work, it always corresponds to sixteen tilt values. Node 6 aggregates the received messages using \cref{eq:mp} to recompute its outgoing messages for the next round, until convergence.}
\end{figure}

In the case of trees, the max-plus procedure is shown to converge to the optimal solutions.
When graphs contain cycles, they only converge to approximately optimal solutions.
Another approach is to use variable elimination to solve \cref{eq:inference} but it is not as scalable.
For a detail treatment of the message passing algorithm we refer the reader to the work of \citeauthor{kok2005maxplus}~\cite{kok2005maxplus}.
The message passing procedure is illustrated in \cref{fig:message-passing} at a specific node from the twenty-seven cells deployment illustrated in \cref{fig:cg}.
We provide an illustration of the message passing procedure in \cref{fig:message-passing} that shows the messages after convergence for our network optimization use case.

Learning the value function of an edge $(i,j)$ can be considered as a standalone, value-based, off-policy RL problem involving two agents.
The equivalent of the state and action spaces for a pair of agent is $(s_i, s_j)$, and $(a_i, a_j$).

\section{More on the Training Performance}

\begin{figure}
	\centering
	\input{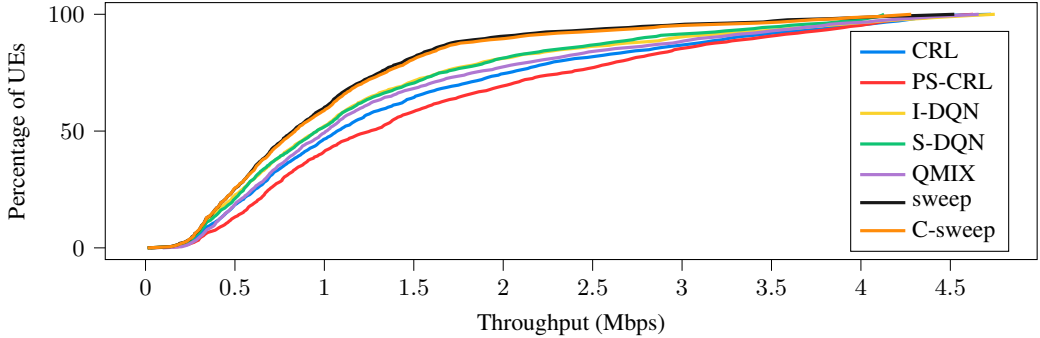}
	\caption{Cumulative density function plot of the throughput for the deployment with \num{27} cells. The more to the right the better}
	\label{fig:cdf}
\end{figure}

Another metric to look at is the throughput CDF as illustrated in \cref{fig:cdf}.
It shows the distribution of throughput across the users.
For the tail of the distribution (below \SI{10}{\percent}), the algorithms have similar performance.
For \SI{20}{\percent} of the users and above, CRL and PS-CRL significantly improve the throughput compared to other methods.
We can also see that the sweeping techniques are less performant than RL based methods.
An explanation is that they cannot adapt well to different user distributions.

\begin{figure}
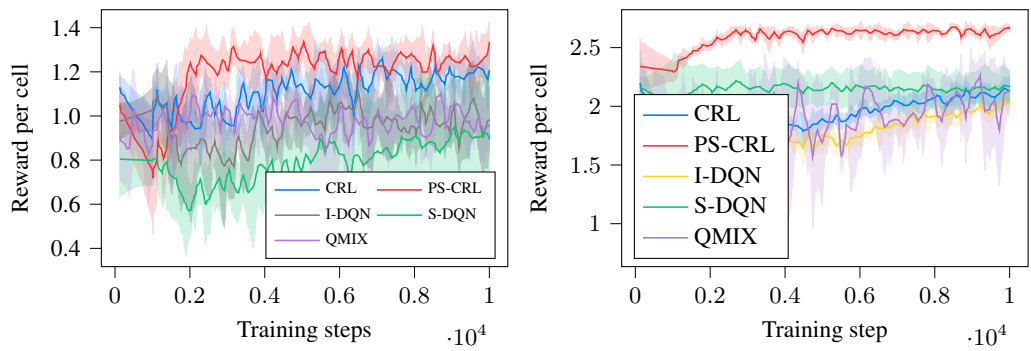

	\centering
	\input{figures/training_performance_27sectors.tex}
	\input{figures/training_performance_207sectors.tex}
	\caption{Training performance on two deployments one with 27 cells (left) and 207 cells (right). Our approach is CRL and PS-CRL, the other algorithms are common multi-agent RL techniques.}
	\label{fig:training}
\end{figure}

The training performance on two different deployment is shown in \cref{fig:training}.
On the larger environment with \num{207} cells, PS-CRL outperforms all the methods by a large margin, it shows better sample efficiency and less variance as well.
A notable fact is that the CRL method seems to still be improving at the end of the training.
I-DQN and QMIX also seems to be still improving.
They show more instability throughout the training.
We suspect that a larger training budget would have yielded better performance for CRL.
However, for those large scenarios, the computational burden of the simulation is very high.
The benefit of parameter sharing is also seen when comparing I-DQN and S-DQN.

\end{document}